\documentclass[conference]{IEEEtran}
\IEEEoverridecommandlockouts
\usepackage{cite}
\usepackage{amsmath,amssymb,amsfonts}
\usepackage{algorithmic}
\usepackage{graphicx}
\usepackage{textcomp}
\usepackage{caption}
\usepackage{romannum}
\usepackage{xcolor}
\def\BibTeX{{\rm B\kern-.05em{\sc i\kern-.025em b}\kern-.08em
    T\kern-.1667em\lower.7ex\hbox{E}\kern-.125emX}}
\begin{document}
\title{Visual Relationship Detection with Language prior and Softmax}
\author{\IEEEauthorblockN{Jaewon Jung}
\IEEEauthorblockA{\textit{Visual Intelligence Research Group} \\
\textit{University of Science and Technology, ETRI SCHOOL}\\
Daejeon, Republic of Korea \\
woodcook486@naver.com}
\and
\IEEEauthorblockN{Jongyoul Park}
\IEEEauthorblockA{\textit{Visual Intelligence Research Group} \\
\textit{ETRI}\\
Daejeon, Republic of Korea \\
jongyoul@etri.re.kr}
}

\maketitle

\begin{abstract}
Visual relationship detection is an intermediate image understanding task that detects two objects and classifies a predicate that explains the relationship between two objects in an image. The three components are linguistically and visually correlated (e.g. \textquotedblleft wear\textquotedblright\space is related to  \textquotedblleft  person\textquotedblright\space and \textquotedblleft shirt\textquotedblright, while  \textquotedblleft laptop\textquotedblright\space is related to \textquotedblleft table\textquotedblright\space and \textquotedblleft on\textquotedblright) thus, the solution space is huge because there are many possible cases between them. Language and visual modules are exploited and a sophisticated spatial vector is proposed. The models in this work outperformed the state of arts without costly linguistic knowledge distillation from a large text corpus and building complex loss functions. All experiments were only evaluated on Visual Relationship Detection and Visual Genome dataset.
\end{abstract}

\begin{IEEEkeywords}
Visual relationship, Image understanding, Deep learning.
\end{IEEEkeywords}

\section{Introduction}
Understanding images is important in computer vision. In deep learning for computer vision, \textit{object classification\space\cite{vgg16,resnet}, detection\space\cite{faster-rcnn,RCNN,fast}, attribute detection\space\cite{arrtiDet}, segmentation\space\cite{maskrcnn,fcn}} and other tasks have improved performance for image understanding. Although these works are still insufficient for understanding images, there is room for improving their performance. Researchers changed their focus to \textit{Scene graph\space\cite{scene_graph}, image captioning\cite{image_cap1}, image retrieve\space\cite{image_retrieval1} } and other related works. \par
One area is \textit{visual relationship detection \cite{lu,visualphrase,Vtrans,vip_cnn,drnet,yu}.} Visual relationships are a type of relationship between objects in an image and consist of subject, predicate, and object; e.g. \{person, ride, motorcycle\}, \{person, eat, hamburger\}, and \{cup, on, table\}. These can be considered sentences without adjectives, adverbs, or the in/definite article. The subject and object in a visual relationship are exactly the same as a subject and object in a sentence. A predicate in a visual relationship is different from a predicate in a sentence. In the dictionary, the meaning of \textquotedblleft predicate\textquotedblright\space is the part of the sentence that contains the verb and gives information about the subject, but the predicate of a visual relationship is similar to a verb. It can be a regular verb, a preposition, a comparative, a prepositional verb, a phrasal verb or other words that could explain a connection between objects. \par 
 One previous approach\space\cite{visualphrase} considered each visual relationship as a one of a class. e.g. \{person, ride, motorcycle\}, \{person, ride, bicycle\} and \{person, ride, skateboard\} are of different classes. This fashion requires numerous data because all possible combinations—meaning the number of predicates times the number of objects squared—are different classes and it results in a huge solution space. Other previous approaches\space\cite{lu,Vtrans,drnet,yu} consider detecting objects (subject and object) and a predicate separately. This fashion reduces the solution space rather than the above approach\space\cite{visualphrase} because solution space of the objects (subject and object) and predicate are decoupled; an object detector and a predicate classifier are only needed in this case but this way still requires large amounts of data. This work follows the later approach to solve the problem. \par
 There are three major difficulties for visual relationship detection: first, intra-class variance; a predicate can be involved with any subject and object. e.g. \{person, eat, pizza\}, \{elephant, eat, grass\}, \{person, use, phone\}, \{person, use, knife\} and so on. These visual relationships are totally difference visually and make the solution space huge. The second difficulty is long-tail distribution. Some of the predicates may occur many times but other certain predicates may only occur once or twice throughout the whole dataset and most of the visual relationships are insufficient for training. This phenomenon brings out a biased dataset and model training result. The third difficulty is class overlapping; some of the predicates in the dataset are almost similar meaning but each data belongs to different class even though their annotations are nearly the same: (near, adjacent, around), (below, under), (look, watch), (next to,feed) etc. \par
 This work utilizes a pair of word vectors, a spatial vector and a union box of two objects’ boxes to detect visual relationships in an image using a language and visual module. The proposed models significantly outperform the state of arts. All experiments in this paper are conducted on the VRD\space\cite{lu}\footnote{VRD dataset link : https://cs.stanford.edu/people/ranjaykrishna/vrd/} and Visual Genome dataset (VG)\space\cite{VG}\footnote{VG dataset link : https://visualgenome.org}.

\section{Related Work}
Object classification\space\cite{vgg16,resnet} is the basis of image understanding, is based on a Convolution Neural Network (CNN). This network learns the features of objects in images and classifies what objects are in an image. As a result of this research field, several CNNs such as VGG16\space\cite{vgg16}, ResNet\space\cite{resnet} called the backbone network outperform object classification. In visual relationship detection, most papers\cite{lu,Vtrans,vip_cnn,drnet,yu}  use these networks to classify the predicate between two objects; this paper employs VGG16\space\cite{vgg16}.\par
Object detection\space\cite{faster-rcnn,RCNN,fast} is the next level of object classification for image understanding. This field also achieves massive success through deep learning. The object detection network localizes objects as bounding boxes in images. R-CNN and Fast/Faster R-CNN\space\cite{faster-rcnn,RCNN,fast} are common object detectors that follow the two-stage approach in which object candidates are proposed while working with RPN\space\cite{faster-rcnn}  and then classify what object is in candidates. Some papers\space\cite{Vtrans,vip_cnn} about visual relationships utilize RPN; they show how the employment of RPN and object detection results are improved. This work employs faster R-CNN\space\cite{faster-rcnn} based on VGG16\space\cite{vgg16}. \par
Human–object interaction recognition\space\cite{HOI1,HOI2} is a subset of the visual relationship. In contrast to visual relationships, a subject is fixed as a person; this field focuses on the interaction between a person and an object or another person. Average Precision (AP) is an evaluation metric of this research field. Specifically, they evaluate the AP of the triplet \{person, verb, object\}, which is called the role AP. Moreover, Ref. \cite{HAction1} is focusing on the action or pose without interaction in an image.\par
 Image captioning is an interesting field in visual tasks in which an image is given as an input and the output is a description that explains that image; this field involves natural language. Recurrent Neural Network (RNN) and Long Short-term Memory (LSTM)\space\cite{LSTM} are used with CNN. Vinyals et al.\space\cite{imagecaption} proposed an architecture in which CNN encodes visual features in an image and RNN decodes it to natural language. \par
 The scene graph\space\cite{scene_graph,Scenegrpah1,Scenegrpah2} is a higher level image understanding. It is a kind of graph in computer science grounded by the visual. Nodes are objects, relationships are edges, and attributes are a sub-node coupled with objects in an image. This field is related to natural language, so some papers\space\cite{Scenegrpah1,ScenegraphbasedDescrip} have attempted to generate a scene graph-based image description. \par
 Visual relationship detection is a superset of human–object interaction. Differently, relationships between any two objects are focused in an image. Some papers\space\cite{lu,visualphrase,Vtrans,vip_cnn,drnet,yu,VRL,Lcue} do work to detect visual relationships; Lu et al.\space\cite{lu} established a visual relationship detection task and introduced a VRD dataset\space\cite{lu} which contains four categories such as verb, preposition, spatial and comparative, and only one predicate exists in a visual relationship regardless of the category. Yu et al.\space\cite{yu} improved the detection performance by expensive linguistic knowledge distillation from an internal and external text corpus. Li et al.\space\cite{vip_cnn} proposed a top-down pipeline. Unlike other approaches\space\cite{lu,Vtrans,drnet,yu}, the visual relationships including subject, predicate, and object are detected simultaneously with RPN\space\cite{faster-rcnn} and the phrase-guided message passing structure (PMPS). Zhang et al.\space\cite{Vtrans} built an equation to embed the visual relationship into space with a class indicator, a location vector, and a visual feature. Ref. \cite{Vtrans,vip_cnn} employed RPN in their architecture and said that cooperating with it improved the object detection result. Liang et al.\space\cite{VRL} proposed a novel framework called deep Variation-structured Reinforcement Learning (VRL) to detect both visual relationships and attributes to understand the global context in an image and use prior language to build a directed semantic graph. Plummer et al.\space\cite{Lcue} conjugated visual and language cues for the localization and grounding of phrases in images and gave a special attention to relationships between people and body parts or clothing. Bo et al.\space\cite{drnet} represented the predicate by using a union box that included the subject, object, and a spatial module consisting of several convolution layers, and detected visual relationships using a deep relational network.
\section{Difficulties of Visual Relationship Detection}
\subsection{Intra-class Variance}
\begin{figure}[h]
\vspace*{-0mm}
\centering
        \includegraphics[width=0.5\textwidth ,height=5cm,keepaspectratio]{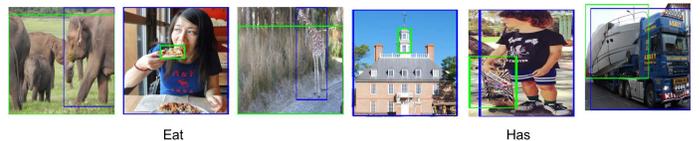}
    \caption{The examples of Intra-class Variance}
    \label{fig:word_embedding_space}
\end{figure}
The approach that detects objects and the predicate independently requires a predicate classifier. A predicate is involved with many subjects and objects. Therefore visual appearance can have a big gap between visual relationships on the same predicate.  \{person, eat, pizza\} and  \{elephant, eat, grass\}  are examples of this.
\subsection{Long Tail Distribution}
This common problem is mentioned in most papers\space\cite{lu,visualphrase,Vtrans,vip_cnn,drnet,yu} and has two aspects: the first is the number of predicates in a dataset and the second is the number of visual relationships. In the VRD\space\cite{lu} and VG\space\cite{VG} dataset, the number of predicate \textquotedblleft on\textquotedblright\space is a huge part of the dataset, but the number of  \textquotedblleft feed\textquotedblright\space and \textquotedblleft talk\textquotedblright\space make up a small part of the dataset. Most of the data are small to train because gathering data and annotation are difficult and expensive; subject, predicate, and object can be obtained easy individually but rarely appear together in an image: \textquotedblleft airplane,\textquotedblright\space \textquotedblleft next to,\textquotedblright\space and \textquotedblleft bag\textquotedblright\space are easily obtained individually, but \{airplane, next to, bag\} is rare.
\subsection{Class Overlapping}
\begin{figure}[h]
\vspace*{-3mm}
\centering
        \includegraphics[width=0.5\textwidth ,height=3cm,keepaspectratio]{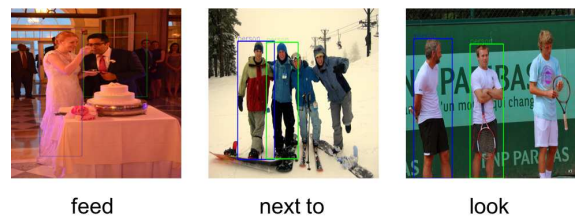}
    \caption{The examples of Class Overlapping}
    \label{fig:word_embedding_space}
\end{figure}
In a predicate list of both datasets, there are some predicates that mean nearly same thing like \textquotedblleft near\textquotedblright\space and \textquotedblleft next to\textquotedblright . According to the dictionary, the difference in the literal meaning between them is vague. For two other cases, some predicates are a superset of others or a subset of others. In visual relationship detection, only one predicate exists between two objects regardless of category. This means that an unrelated predicate that has a totally different meaning can be chosen when the feature vector is close. These phenomena cause wrong classification results.
\section{Approach}

\begin{figure*}[htbp]
\centering
        \includegraphics[width=\textwidth ,height=15cm,keepaspectratio]{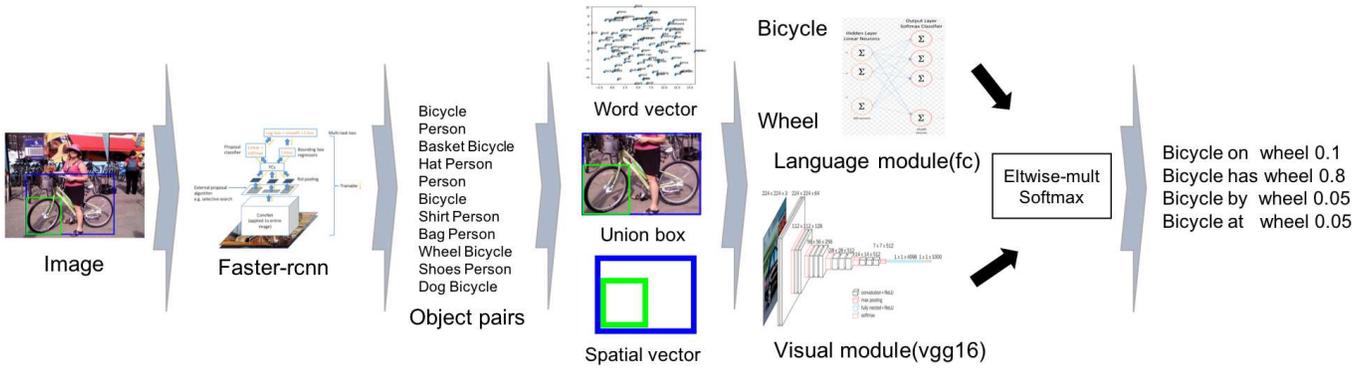}
    \caption{An overview of a full visual relationship detection model. An image is given as an input. Faster R-CNN\cite{faster-rcnn} based on VGG16\cite{vgg16} detects object(s) and pairs of objects called candidates are generated from detected objects. Each pair of objects is fed to the predicate classifier with a pair of word vectors, a spatial vector, and a union box that includes subject and object. The language and visual module produce results, multiply it element-wise and softmax is applied to the result. }
    \label{fig:pipepline}
    \vspace*{-4mm}
\end{figure*}
\subsection{Visual Module}
A visual module is VGG16\space\cite{vgg16} and trained similar to object fine-tuning with a softmax loss for classifying the predicate using a union box that includes two objects in an image as an input. \par
This work relieves ambiguous inferences using only the union box containing two objects, a pair of word vectors, and/or a spatial vector are/is used together. Therefore, variant visual modules are newly created based on the visual module such as a spatial and visual module (SV), a visual and word vector module (VW), and a spatial, visual, and word vector module (SVW).
\subsection{Language Module}
\begin{equation} \label{eq1_language_module}
\begin{split}
W \times [wordvector(subject),wordvector(object)] + b
\end{split}
\end{equation}
A language module is trained with the softmax loss instead of the K, L loss in \cite{lu}, and takes a pair of subject and object word vectors which is 600 dimensions as an input. These are fed to a fully connected layer and produce 70 dimensions vector as an output and 70 is the number of predicates in the dataset. W is 70 $\times$ 600 and b is 70 dimensions in (1). This simple training approach fulfils the K and, L loss. In \cite{lu}, the L loss gives a higher likelihood to high-frequency data and a lower likelihood to low-frequency data in a training dataset. The K loss enforces similar visual relationships getting close, and far away from dissimilar visual relationships; e.g. \{person, eat, pizza\}, \{person, eat, hamburger\} are similar and  \{car, has, wheel\} is dissimilar from them.  It means the language module in \cite{lu} produces similar likelihood when visual relationship are close. Without the L loss, the language module in this work naturally assigns the appropriate likelihood to predicates depending on the frequency because several predicates can exist when subject and object are given, only one predicate is annotated for a visual relationship and loss is the softmax ; for an example, \textquotedblleft wear\textquotedblright\space has a higher likelihood than \textquotedblleft hold\textquotedblright\space when the subject is \textquotedblleft  person\textquotedblright\space and the object is \textquotedblleft shirt\textquotedblright . Without the K loss, the property of word vector leads similar visual relationships to get close and further away from dissimilar visual relationships. \{person, ride, bicycle\} and \{person, ride, motorcycle\} naturally get close because \textquotedblleft bicycle\textquotedblright\space and  \textquotedblleft motorcycle\textquotedblright\space word vectors are close in a word vector space. This means that the language module in this work produces nearly same likelihood for \textquotedblleft ride\textquotedblright\space when the subject is \textquotedblleft person\textquotedblright\space and the object is \textquotedblleft bicycle\textquotedblright\space or \textquotedblleft motorcycle\textquotedblright . \par
As with the visual module, a spatial vector is concatenated on the pair of word vectors before the fully connected layer as a new module called the language and spatial module (LS) to relieve the ambiguous inference based on only the pair of word vectors.
\subsection{Spatial Vector}
\begin{equation} \label{eq2_spatial_module}
\begin{split}
    [ IOU, x,y,S_{subject}/S_{image},S_{object}/S_{image},\\cflag_{subject},cflag_{object}]
\end{split}
\end{equation}
 This work proposes a sophisticated spatial vector different from \cite{yu}. The spatial vector in \cite{yu} only reflects only each object’s bounding box normalized location and size in an image. This encoding is insufficient to classify predicates because predicates do not depend on location in an image. The proposed vector encodes the intersection over union (IOU) and normalized relative location (x, y) based on the subject box center, normalized subject and object size, and contain flag (cflag) for subject and object; cflag for a subject is 1 if the subject box contains the object box, and 0 otherwise, and vice versa. $S_{subject,object}$ is the size of the bounding box for each and $S_{image}$ is the size of an image in (2).
\subsection{Model Variants}
\begin{equation} \label{eq3_combined_model_loss}
\begin{split}
softmax(visaul module \times language module)
\end{split}
\end{equation}
Several components are available in the modules including a spatial vector, word vectors and a union of the bounding box. The model consists of two modules: the language and visual module. The base component of a language module is a pair of word vectors and spatial vector can be added to the language module. The base component of a visual module is a union box, and word vector and/or spatial vector can be added to the visual module. Furthermore, the language and visual modules can be trained together or separately. Therefore, possible models are L, LS, V, VW, SVW, SV, L + V, L + VW, L + SV, L + SVW, LS + V, LS + VW, LS + SV, and LS + SVW for an experiment (\textquotedblleft +\textquotedblright\space means that two modules are combined). When the language and visual modules are jointly trained, the loss function is (3). \textquotedblleft$\times$\textquotedblright\space means element-wise multiplication. Each module produces 70 dimensions vector. 
\section{Experiments}
In this section, predicate prediction, phrase, and relationship detection are conducted on the VRD and VG datasets\space\cite{lu,VG}. In predicate prediction, the models take an image and set of localized subjects and objects as an input and predict the set of possible predicates between pairs of objects. In phrase detection, the models take an image as an input, detect the phrase for a set of visual relationships and localize the entire relationship as one bounding box that has at least a 0.5 overlap with the ground truth box. In relationship detection, the models take an image as an input, detect the relationship as a set of visual relationships and localize the subject and object in the image that have at least a 0.5 overlap with their ground truth boxes at the same time.
\subsection{Dataset}
Particularly for the VG dataset\space\cite{VG}, previous works cleaned up the VG dataset on their way. For a fair comparison, the visual relationships that are consistent with the VRD dataset\space\cite{lu} are extracted from the VG dataset to compare with Yu’s results. For a zero-shot test, a dataset must consist of unseen visual relationships that never occurred in the training dataset for the VRD and VG datasets. The task that extracts unseen visual relationships from the original is applied on both original datasets.
\subsection{Evaluation Metric}
Then, Recall@n (R@n) is chosen as a metric becasue it is used in \cite{lu,yu} and the evaluation algorithm is modified based on \cite{lu}. Additionally, the evaluation fashion from \cite{yu} is applied. Consistent with \cite{yu}, the number of chosen predictions (k) per object pair is hyper-parameter and shows R@n for different k for fair and equal comparison.
\subsection{Predicate Prediction}
\begin{table}[ht]
 	\centering
 	 \captionsetup{position=bottom}
 	  \renewcommand{\arraystretch}{1.5}
 	  \LARGE
 	 \caption{Predicate prediction on the VRD dataset. In \cite{yu}, \textquotedblleft U\textquotedblright\space is the union box that includes two objects, \textquotedblleft SF\textquotedblright\space is the spatial vector in their work,  \textquotedblleft W\textquotedblright\space is the word-embedding-based semantic representations, \textquotedblleft L\textquotedblright\space is the linguistic knowledge distillation, \textquotedblleft S\textquotedblright\space is the student network, \textquotedblleft T\textquotedblright\space is the teacher network and \textquotedblleft S+T\textquotedblright\space is the combination of two networks. In this work, \textquotedblleft L\textquotedblright\space is a language module that uses word vectors, \textquotedblleft S\textquotedblright\space is the proposed spatial vector, \textquotedblleft V\textquotedblright\space is the same as \textquotedblleft U\textquotedblright\space, \textquotedblleft W\textquotedblright\space is the word vectors in the visual module, \textquotedblleft +\textquotedblright\space means that the two modules placed before and after the \textquotedblleft +\textquotedblright\space are used  together. Before the double vertical line is the general performance and after is the zero-shot performance}
 	  \resizebox{\columnwidth}{!}{
 	\begin{tabular}{c|c|c|c|c||c|c|c|c}
 		\hline  &  R@50 & R@100 & R@50& R@100 & R@50 & R@100 & R@50 & R@100  \\
 		&  k=1 & k=1 &  k=70 & k=70 & k=1 & k=1 & k=70 & k=70 \\
 		%\multicolumn{2}{c}{} \\[3pt]
 		\hline  VRD \cite{lu} & 47.87 & 47.87  & - & - & 8.45 & 8.45  & - & -\\  
 		\hline  U+W+SF \cite{yu} & 41.33 & 41.33 & 72.29 & 84.89 & 14.13 & 14.13  & 48.13 & 69.41\\ 
 		\hline  U+W+L:S \cite{yu} & 42.98 & 42.98  & 71.83 & 84.94 & 13.89 & 13.89  & 51.37 & 72.53\\ 
 		\hline  U+W+L:T \cite{yu} & 52.96 & 52.96  & 83.26 & 88.98 & 7.81 & 7.81  & 32.62 & 40.15\\ 
 		\hline  U+SF+L:S \cite{yu} & 41.06 & 41.06  & 71.27 & 84.81 & 14.33 & 14.33  & 48.32 & 69.01\\ 
 		\hline  U+SF+L:T \cite{yu} & 51.67 & 51.67  & 83.84 & 87.71 & 8.05 & 8.05  & 32.77 & 41.51\\ 
 		\hline  U+W+SF+L:S \cite{yu} & 47.50 & 47.50  & 74.98 & 86.97 & 16.98 & 16.98  & 54.20 & 74.65\\ 
 		\hline  U+W+SF+L:T \cite{yu} & 54.13 & 54.13  & 82.54 & 89.41 & 8.80 & 8.80  & 32.81 & 41.53\\ 
 		\hline  U+W+SF+L:T + S \cite{yu} & 55.16 & 55.16  & 85.64 & 94.65 & - & -  & - & -\\ 
 		\hline  L & 44.09  &44.09 & 75.48 & 86.69 & 10.86 & 10.86 & 50.55 & 69.71\\  
 		\hline  LS & 48.19 & 48.19 & 78.31 & 88.40 & 15.82 & 15.82  & 55.09 & 74.85\\   
 		\hline  SVW & 48.57 & 48.57 & 78.04  & 88.30  & 16.85  & 16.85  & 55.77 & 74.85 \\ 
 		\hline  L+V & 49.77 & 49.77 & 79.99  & 88.81  & 14.88  & 14.88  & 54.40 & 72.51 \\   
 		\hline  LS+VW & 53.05 & 53.05  &  85.12 & 93.17 & 20.78 & 20.78 & 64.67 & 81.35 \\ 
 		\hline  LS+SV & 53.37 & 53.37  &  85.61 & 93.74 & 21.21 & 21.21 & 65.78 & 82.37\\ 
 		\hline  LS+SVW & 55.16 & 55.16  & 88.88  & 95.18  & 21.38 & 21.38  & 64.49 & 83.49 \\
 		\hline
 	\end{tabular}}
 	\label{tab:table_predicate_prediction} 
\end{table}
Table  \Romannum{1} shows the results of predicate prediction on the VRD dataset\space\cite{lu}. The results of L model that only considers a pair of word vector is 44.09 R@50,100 when k = 1. The most referenced predicates are \textquotedblleft on\textquotedblright\space and \textquotedblleft  wear\textquotedblright, as these are common predicates in the dataset. Better performance is produced from the LS model which takes a pair of word vectors and spatial vector. \textquotedblleft  on\textquotedblright\space and \textquotedblleft wear\textquotedblright\space are the most commonly referenced but this model can distinguish spatial predicates. The L + V model outperforms the model in \cite{lu} for same condition that a pair of word vectors and a union box are used. In particular, the zero-shot result is nearly 6\% higher than \cite{lu}. The reasion is that the language prior is well obtained in the model in this work rather than the model in \cite{lu}. The result from SVW model is a huge improvement over U + W + SF, U + W + L:S and U + SF + L:S. This means that coupling a spatial vector and a pair of word vectors on the visual module works better than \cite{yu} and shows the possibility that a model can perform better without linguistic knowledge distillation. Next, the language and visual modules are jointly trained, and LS + SV, LS + VW and LS + SVW are models. The last model outperforms U + W + SF + L:T (+ S) without linguistic knowledge distillation. Despite putting more information into models, those models’ performances are nearly same. A verb predicate is considered another category predicate such as a preposition or a spatial when k = 1 because of class overlapping. For an example, models predict \textquotedblleft on\textquotedblright\space instead of \textquotedblleft run\textquotedblright .

\begin{table}[ht]
 	\centering
 	\vspace*{-10pt}
 	 \captionsetup{position=bottom}
 	 \LARGE
 	 \caption{Predicate prediction on VG dataset. The notations are same as in Table \Romannum{1}}
 	  \resizebox{\columnwidth}{!}{
 	\begin{tabular}{c|c|c|c|c||c|c|c|c}
 		\hline  &  R@50 & R@100 & R@50& R@100 & R@50 & R@100 & R@50 & R@100  \\
 		&  k=1 & k=1 &  k=70 & k=70 & k=1 & k=1 & k=70 & k=70 \\
 		%\multicolumn{2}{c}{} \\[3pt]
 		\hline  U+W+SF+L:S & 49.88 & 49.88  & 88.14 & 91.25  & 11.28  & 11.28  & 72.96 & 88.23 \\  
 		\hline  U+W+SF+L:T & 55.02 & 55.02  & 91.47  & 94.92  & 3.94 & 3.94  & 47.62 & 62.99 \\   
 		\hline  U+W+SF+L:T+S & 55.89 & 55.89  & 92.31 & 95.68 & - & -  & - & - \\ 
 		\hline  SVW & 65.59  & 65.73 & 96.37 & 98.90 & 16.82 & 16.82 & 86.33 & 95.02\\  
 		\hline  LS+SVW & 70.99 & 71.12 & 97.98 & 99.37 & 19.68 & 19.68  & 89.00 & 95.72\\  
 		\hline
 	\end{tabular}}
 	\label{tab:table_predicate_prediction_vg} 
 	\vspace*{-0mm}
\end{table}
Table \Romannum{2} shows the predicate prediction result on the VG dataset\space\cite{VG}. The two models in this work unquestionably outperform Yu’s model\space\cite{yu}. SVW model surpasses Yu’s model without the language module. The dataset in \cite{yu} is not shared and the experiment is conducted on the newest visual relationship version 1.4 of VG dataset\cite{VG}. Like \cite{yu}, the images are randomly shuffled and split into training and test set. The visual relationships that match the VRD dataset\space\cite{lu} is extracted from the VG dataset\space\cite{VG}. The dataset that is used in this experiment contains 26,180 images and 71,269 visual relationships for training and 13,092 images and 36,184 visual relationships for testing. The number of unseen relationships that never occur in the training dataset is 2,692.
\subsection{Phrase and Relationship Detection }
\begin{table}[ht]
    \vspace*{-2mm}
 	\centering
 	 \renewcommand{\arraystretch}{2.0}
 	 \captionsetup{position=bottom}
 	 \caption{Phrase detection result on VRD dataset. The notations are same as in Table \Romannum{1}. Above the double horizontal line is the performance to compare the result to \cite{yu} and below is the performance to compare to compare it to \cite{lu}.}
 	 \LARGE
 	\resizebox{\columnwidth}{!}{
 	\begin{tabular}{c|c|c|c|c|c|c||c|c|c|c|c|c}
 	
 		\hline  &  R@50 & R@100 & R@50 & R@100 & R@50 & R@100 & R@50 & R@100 & R@50 & R@100 & R@50 & R@100 \\
 		&  k=1 & k=1 & k=10 & k=10 &  k=70 & k=70 &  k=1 & k=1 & k=10 & k=10 &  k=70 & k=70 \\
 		%\multicolumn{2}{c}{} \\[3pt]
 		\hline  VIP-CNN \cite{vip_cnn} & 22.78 & 27.91 & - & -  & -  & -  & - & - & - & - & - & - \\
 		\hline  VRL \cite{VRL} & 21.37 & 22.60 & - & -  & - & -  & 9.17 & 10.31 & - & - & - & - \\
 		\hline  Linguistic Cues \cite{Lcue} & - & - & 16.89 & 20.70  & - & - & - & -  & 10.86 & 15.23 & - & -  \\
 		\hline  U+W+SF+L:S \cite{yu} & 19.15 & 19.98 & 22.95 & 25.16 & 22.59  & 25.54 & 10.44 & 10.89 & 13.01 & 17.24 & 12.96 & 17.24\\
 		\hline  U+W+SF+L:T \cite{yu} & 22.46 & 23.57  & 25.96 & 29.14 & 25.86 & 29.09 & 6.54 & 6.71& 9.45 & 11.27 & 7.86 & 9.84\\ 
 		\hline  U+W+SF+L:T + S \cite{yu} & 23.14 & 24.04 & 26.47 & 29.76 & 26.32 & 29.43 & - & - & - & - & - & -\\
 		\hline  LS+SV & 32.15 & 33.00 & 41.58 & 49.45 & 41.68 & 49.89 & 12.23 & 12.66 & 22.75 & 32.59 & 23.26 & 34.21\\  
 		\hline \hline  VRD \cite{lu} & 16.17 & 17.03 & -  & -  & -  & -  & 3.36 & 3.75 & - & - & - & - \\ 
 		\hline  LS+SV & 17.00 & 19.03 & 18.94 & 23.01 & 18.95 & 23.06 & 7.35 & 8.12 & 9.83 & 13.08 & 9.92 & 13.43\\  
 		\hline
 	\end{tabular}}
 	\label{tab:table_phrase_detection} 
 	\vspace*{-0mm}
\end{table}
Table \Romannum{3} and \Romannum{4} show the results on phrase and relationship detection. These experiments are conducted on an object detection result and Lu et al.\cite{lu} conducted experiments on an object detection result from RCNN\space\cite{RCNN}. For a fair comparison, models in this work are evaluated on the same result which is shared from \cite{lu}. All models perform slightly better than \cite{lu} on phrase and relationship detection. For comparision to \cite{yu}, faster R-CNN\space\cite{faster-rcnn} based on VGG16\cite{vgg16} is trained for VRD objects for this experiment.
\begin{table}[h]
 	\centering
 	 \captionsetup{position=bottom}
 	 \caption{Relationship detection on VRD dataset. The notations are same as in Table \Romannum{1}. Above the double horizontal line is the performance to compare the result to \cite{yu} and below is the performance to compare to compare it to \cite{lu}.}
 	 \renewcommand{\arraystretch}{1.8}
 	 \LARGE
 	\resizebox{\columnwidth}{!}{
 	\begin{tabular}{c|c|c|c|c|c|c||c|c|c|c|c|c}
 		\hline  &  R@50 & R@100 & R@50 & R@100 & R@50 & R@100 & R@50 & R@100 & R@50 & R@100 & R@50 & R@100 \\
 		&  k=1 & k=1 & k=10 & k=10 &  k=70 & k=70 &  k=1 & k=1 & k=10 & k=10 &  k=70 & k=70 \\
 		%\multicolumn{2}{c}{} \\[3pt]  
 		\hline  VIP-CNN \cite{vip_cnn} & 17.32 & 20.01 & - & -  & -  & -  & - & - & - & - & - & - \\
 		\hline  VTRANS \cite{Vtrans} & 14.07 & 15.20 & - & -  & - & -  & 1.71  & 2.14 & - & - & - & -  \\
 		\hline  VRL \cite{VRL} & 18.19& 20.79 & - & -  & - & -  & 7.94 & 8.52& - & - & - & -  \\
 		\hline  Linguistic Cues \cite{Lcue} & - & - & 15.08 & 18.37  & - & -  & - & - & 9.67 & 13.43 & - & - \\
 		\hline  U+W+SF+L:S \cite{yu} & 16.57 & 17.69 & 19.92 & 27.98 & 20.12  & 28.94 & 8.89 & 9.14 & 12.31 & 16.15 & 12.02 & 15.89\\
 		\hline  U+W+SF+L:T \cite{yu} & 18.56 & 20.61  & 21.91 & 29.41 & 21.98 & 31.13 & 6.07& 6.44 & 7.82 & 9.71 & 8.75 & 10.21\\ 
 		\hline  U+W+SF+L:T + S \cite{yu} & 23.14 & 24.04 & 26.47 & 29.76 & 26.32 & 29.43 & - & - & - & - & - & -\\
 		\hline  LS+SV & 30.25 & 31.06 & 39.49 & 47.38 & 39.60 & 47.80 & 11.97 & 12.40 & 22.07 & 31.73 & 22.58 & 33.36\\ 
 		\hline \hline  VRD \cite{lu} & 13.86 &  14.70 & -  & -  & -  & -  & 3.13 & 3.52 & - & - & - & -\\ 
 		\hline  LS+SV & 15.05 & 16.73 & 16.82 & 20.49 & 16.83 & 20.54 & 6.75 & 7.35 & 8.98 & 11.80 & 9.06 & 12.14\\ 
 		\hline
 	\end{tabular}}
 	\vspace*{-0mm}
 	\label{tab:table_relationship_detection} 
\end{table}

\subsection{ Benefit of Proposed Spatial Vector }
The spatial vector in \cite{yu} is replaced instead of the proposed vector in this model to verify the capacity of the proposed spatial vector. Table  \Romannum{5} shows that the proposed vector improves 2\% and 4\% performance rather than the spatial vector in \cite{yu} on Recall@50 for k = 1. On zero-shot detection, the performance is improved nearly 5\% and 3\% on Recall@50 for k = 1. This shows that proposed one has effect of detecting unseen visual relationships. \par
The proposed vector consists of the relative information except each size of subject and object in an image. IOU means the overlap ratio between two boxes; it does not tell us where the overlap is because it is a scalar value. cflag for a subject and an object, and the relative location can reflect the individuality of a predicate and these complement IOU.
\begin{table}[!htbp]
 	\centering
 	 \captionsetup{position=bottom}
 	 \renewcommand{\arraystretch}{1.5}
 	 \caption{Predicate prediction on VRD dataset to verify proposed spatial vector. The notations are same as in Table \Romannum{1}. \textquotedblleft SF\textquotedblright\space is spatial vector in \cite{yu} }
 	 \resizebox{\columnwidth}{!}{
 	\begin{tabular}{c|c|c|c|c||c|c|c|c}
 		\hline  &  R@50 & R@100 & R@50& R@100 & R@50 & R@100 & R@50 & R@100  \\
 		&  k=1 & k=1 &  k=70 & k=70 & k=1 & k=1 & k=70 & k=70 \\
 		%\multicolumn{2}{c}{} \\[3pt]
 		\hline  (SF)VW & 45.58  & 45.58 & 77.10 & 87.98 & 13.60 & 13.60 & 53.63 & 74.16\\  
 		\hline  L(SF)+(SF)VW & 50.53 & 50.53 & 81.99 & 91.08 & 15.99 & 15.99  & 56.63 & 76.98\\ 
 		\hline
 		\hline  SVW & 48.57 & 48.57 & 78.04  & 88.30  & 16.85  & 16.85  & 55.77 & 74.85 \\  
 		\hline  LS+SVW & 55.16 & 55.16  & 88.88  & 95.18  & 21.38 & 21.38 & 64.49 & 83.49 \\   
 		\hline
 	\end{tabular}}
 	\label{tab:spatialvector_experiment}
\end{table}

\subsection{ Benefit of Word Vector for Zero-shot }
\begin{figure}[h]
\vspace*{-0mm}
\centering
        \includegraphics[width=1.0\textwidth ,height=4cm,keepaspectratio]{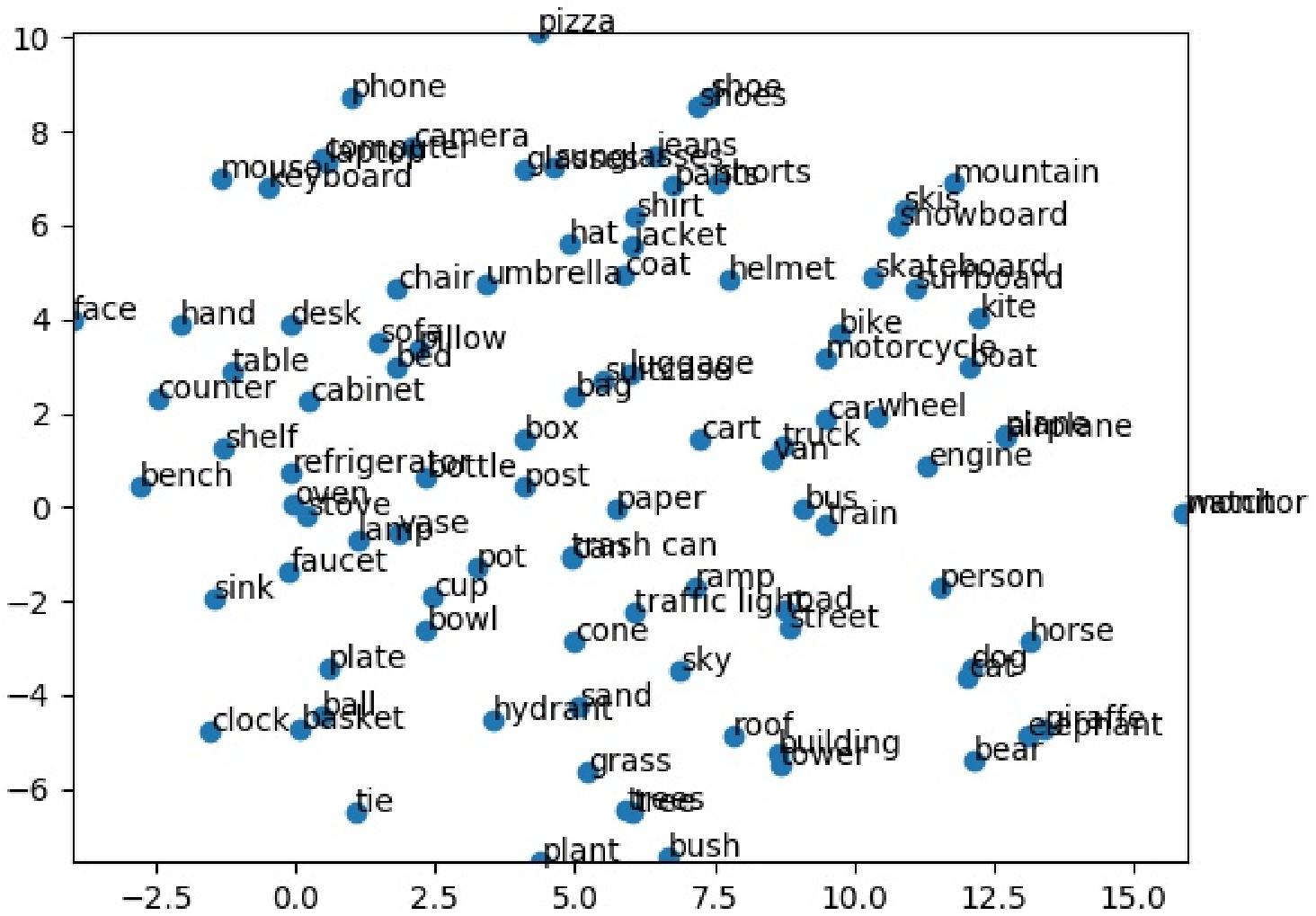}
    \caption{Word vector embedding space}
    \label{fig:word_embedding_space}
\end{figure}
 Word vector embedding space provides clusters that have semantically similar word vectors. These clusters help detect unseen visual relationships that never occurred in training dataset. For example, \textquotedblleft jacket\textquotedblright\space and \textquotedblleft shirt\textquotedblright\space resemble one another with regard to wearing but they are different. One is outer clothing and the other is regular clothing. If \{person, wear, jacket\} only occurs in test dataset, Proposed model easily detects this relationship because \textquotedblleft jacket\textquotedblright\space and \textquotedblleft shirt\textquotedblright\space word vectors are really close in Fig. 4. Particularly when the spatial vector is given, \{person, ride, motorcycle\}, which never occurs in training dataset can be detected easily rather than detection using only word vector. \{person, ride, bicycle\} is semantically and spatially related to \{person, ride, motorcycle\}. From the view of riding a vehicle, the pose is similar between them and the spatial vectors of those are naturally almost the same. The predicate \textquotedblleft ride\textquotedblright\space can be detected with high confidence between \textquotedblleft person\textquotedblright\space and \textquotedblleft motorcycle\textquotedblright .

\section{Conclusions}
The main contribution is outperformed result which can be obtained by simple modification on \cite{lu}. The proposed spatial vector is better than the spatial vector in \cite{yu} on visual relationship detection.  Especially zero-shot performance is significantly improved using proposed spatial vector and word vectors. This paper mentions class overlapping for the first time, which is difficult in visual relationship detection. This work will be shared in public.

\section*{Acknowledgment}
This work was supported by Electronics and Telecommunications Research Institute (ETRI) grant funded by the Korean government. [18ZS1100, Core Technology Research for Self-Improving Artificial Intelligence System]
\bibliographystyle{IEEEtran}
\bibliography{egbib}
\end{document}